\documentclass[lettersize,journal]{IEEEtran}
\usepackage{amsmath,amsfonts}
\usepackage{array}
\usepackage[caption=false,font=normalsize,labelfont=sf,textfont=sf]{subfig}
\usepackage{textcomp}
\usepackage{stfloats}
\usepackage{url}
\usepackage{verbatim}
\usepackage{graphicx}
\usepackage{cite}
\usepackage{amssymb} 
\usepackage{bm}
\usepackage{makecell}
\usepackage{bbding}

\usepackage{graphics} 
\usepackage{epsfig} 

\usepackage{mathtools}
\usepackage{enumitem}
\usepackage{times} 

\usepackage[utf8]{inputenc} 
\usepackage[T1]{fontenc} 
\usepackage{hyperref} 
            
\usepackage{booktabs} 
\usepackage{nicefrac} 
\usepackage{microtype} 
\usepackage{xcolor} 
\usepackage{cleveref}
\usepackage{float}
\usepackage{multirow}
\usepackage{diagbox}

\usepackage{algorithm}
\usepackage{algpseudocode}
\usepackage{arydshln}
\usepackage{marvosym}

\usepackage{listings}
\usepackage{siunitx} 
\usepackage{capt-of}
\usepackage{cuted}

\begin{document}
\newcommand{\ourname}{\texttt{DREAM}}
\bstctlcite{IEEEexample:BSTcontrol}

\title{\ourname{}: Dynamic Resilient Spatio-Semantic Memory \\ with Hybrid Localization for Mobile Manipulation}

\author{Zhijie Yan, Shufei Li, Ze Zhang, Xin Liu, Yuhang Zheng, Zuoxu Wang%
\thanks{Corresponding author: Zuoxu Wang, e-mail: zuoxu\_wang@buaa.edu.cn.}%
\thanks{$^{1}$Zhijie Yan, Ze Zhang, Xin Liu and Zuoxu Wang are with School of Mechanical Engineering and Automation, Beihang University, Beijing 100191, China.}%
\thanks{$^{2}$Shufei Li is with Department of Systems Engineering, City University of Hong Kong, Hong Kong SAR 518057, China.}%
\thanks{$^{3}$Yuhang Zheng is with School of Computing, National University of Singapore,
21 Lower Kent Ridge Rd 119077, Singapore.}%

}
\maketitle

\begin{strip}
    \vspace{-33mm}
    \centering
    
    \includegraphics[width=1\linewidth]{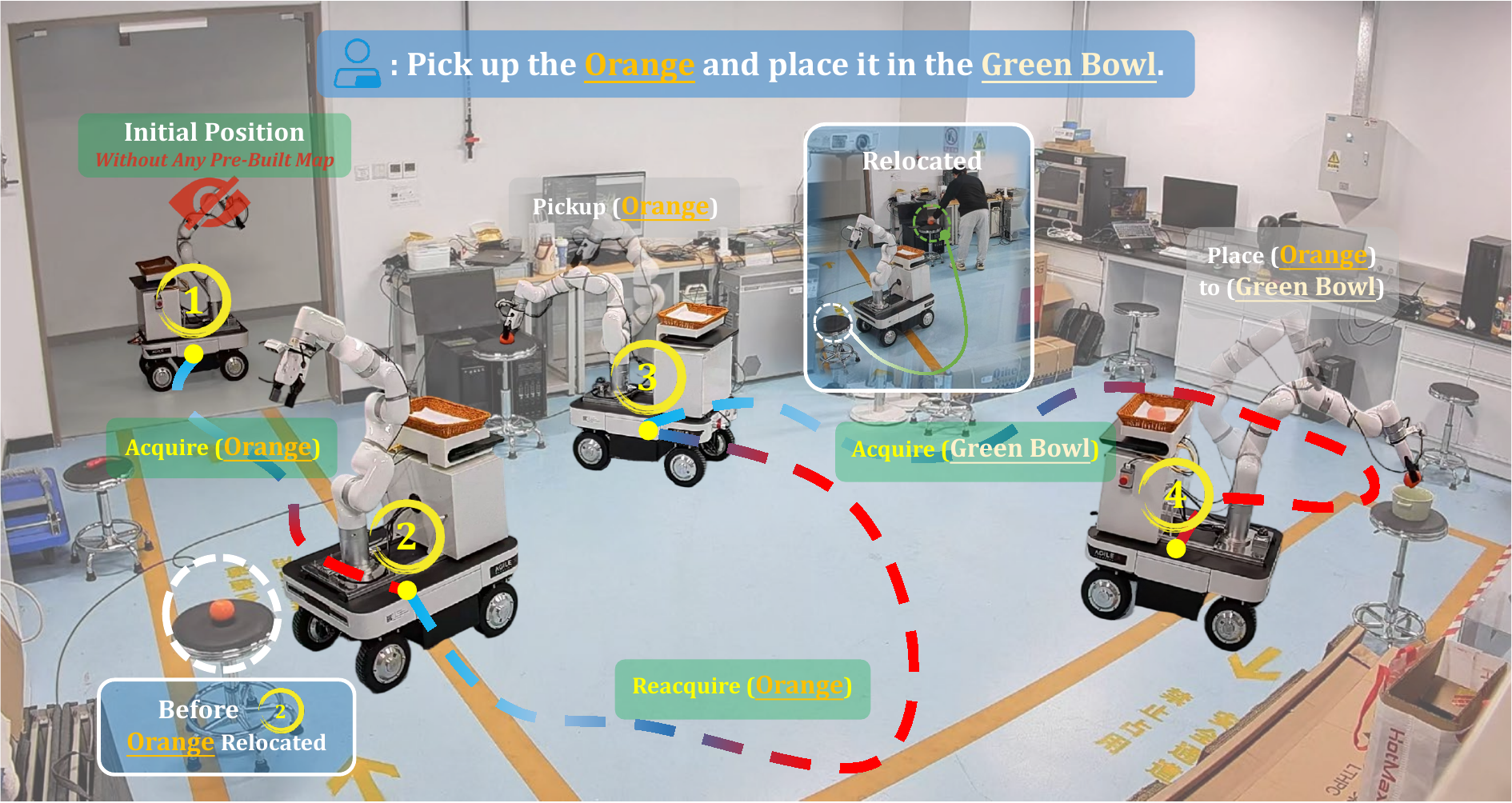}
    \vspace{-7mm}
    \captionof{figure}{\textbf{Overview of the \ourname{} framework.} \ourname{} is a mobile manipulation framework for dynamic, previously unseen indoor environments without any pre-built map. Given a natural-language instruction, the robot actively explores the scene to acquire the target object. When human intervention relocates the object, \ourname{} leverages hybrid localization and a dynamic spatio-semantic memory to reacquire the target, continuously updating task-relevant scene entities and replanning its actions. This enables consistent scene understanding and reliable mobile manipulation under frequent environment changes.}
    \label{fig:teaser}
    \vspace{-4mm}
\end{strip}

\begin{abstract}
    Reliable mobile manipulation in dynamic indoor environments requires a scene representation that remains geometrically consistent, semantically queryable, and computationally bounded as the world changes.
    Existing systems often rely on pre-built maps, static-scene assumptions, or highly accurate camera poses; when target objects are relocated or pose estimates are corrected, navigation and manipulation may be driven by stale or misaligned scene information.
    This paper presents \ourname{}, a real-robot mobile manipulation framework that couples perception, memory, localization, navigation, and manipulation for previously unseen indoor environments without a pre-built map.
    \ourname{} builds an online spatio-semantic voxel memory from RGB-D observations registered by a LiDAR-inertial-visual SLAM backend, uses pose-graph-aware Redundancy-Aware Memory Pruning (RMP) to update historical observations under pose corrections while keeping long-horizon observation history bounded, and combines language-conditioned 3D retrieval, open-vocabulary image detection, and multimodal large-language-model (mLLM)-based verification for target localization and reacquisition.
    Real-robot experiments in four dynamic indoor laboratory scenes show that \ourname{} improves long-horizon task success from $40\%$--$60\%$ to $55\%$--$70\%$ compared with DynaMem, while maintaining a $0.37$--$0.63$\,GB memory footprint and a $0.43$--$0.53$\,s online memory-update time across scenes.
    Code, CAD model, and demonstrations are available at: \href{https://bjhyzj.github.io/dream-web}{https://bjhyzj.github.io/dream-web}.
    \end{abstract}

    \begin{IEEEkeywords}
    Spatio-Semantic Memory, Dynamic Environments, Hybrid Localization, Mobile Manipulation.
    \end{IEEEkeywords}
    
    \section{Introduction}
    \IEEEPARstart{M}{obile manipulation} in real-world settings requires a closed loop over perception, spatio-semantic memory, localization, navigation, and manipulation. Human activity and robot actions continuously change object layouts, occlusions, and traversability. Systems that depend on pre-built maps~\cite{conceptgraphs, hovsg}, static-scene assumptions~\cite{ok-robot}, or drift-free camera poses~\cite{dynamem, dovsg} can therefore operate on obsolete scene representations, leading to brittle navigation and manipulation in dynamic environments.
    
    Recent open-vocabulary robotic systems have made substantial progress by combining foundation models with 3D maps, semantic fields, scene graphs, and modular navigation/manipulation pipelines~\cite{ok-robot, conceptgraphs, dovsg, dynamem}. These systems make language-conditioned manipulation more flexible than category-specific or training-scene-specific policies. However, many of them are designed around an offline or pre-built scene representation, or they evaluate in settings where the pose stream and the scene state are assumed to remain sufficiently stable. Such assumptions are restrictive for service robots deployed in previously unseen homes, offices, and laboratories, where objects may be moved after being observed, temporary occlusions can invalidate earlier detections, and the robot must continue operating while its SLAM pose graph is being refined.
    
    This paper considers previously unseen indoor environments in which no offline pre-built map, manually annotated scene model, or scene-specific prior is available before deployment. The robot must instead observe the environment online and build its internal representation during execution. The central challenge is therefore not only to detect a target object once, but to maintain an actionable scene memory throughout the task. Such a memory should be built online from raw robot observations, remain geometrically consistent when loop closures or pose-graph updates occur, support open-vocabulary target localization and reacquisition after object relocation, and stay compact enough for long-horizon operation.
    
    \ourname{} addresses this setting as a complete real-robot framework for dynamic, previously unseen indoor environments without a pre-built map. It maintains an online spatio-semantic voxel memory from RGB-D observations and SLAM poses, updates this memory when the scene changes, and uses a lightweight RMP mechanism to reintegrate historical observations under pose-graph updates while keeping the stored observation history bounded. On top of this memory, a hybrid localization and reacquisition pipeline combines language-conditioned 3D retrieval, open-vocabulary image detection, and mLLM-based verification. A value-guided exploration and navigation module then selects task-relevant frontiers and manipulation-aware docking poses, while a modular grasping and placement stack completes the pickup/place stages in cluttered scenes.
    
    Unlike methods that primarily study static reconstruction or offline scene querying, \ourname{} is evaluated as a closed-loop robot system. The SLAM backend provides local odometry and global pose-graph consistency; the dynamic spatio-semantic memory stores task-relevant 3D semantic evidence; the hybrid localizer grounds language queries by combining memory retrieval, image-space detection, and semantic verification; and the navigation/manipulation modules convert the recovered target state into executable robot actions. This system-level integration is important because failures in dynamic mobile manipulation often arise from mismatches between modules, for example when a semantic memory is not updated after pose correction, when a moved target is still localized at an obsolete position, or when a successful detection cannot be converted into a reachable manipulation pose.
    The main contributions are summarized as follows:
    \begin{itemize}
        \item A closed-loop real-robot framework is introduced for mobile manipulation in dynamic, previously unseen indoor environments without a pre-built map, integrating SLAM, semantic memory, target reacquisition, navigation, grasping, and placement.
        \item A dynamic spatio-semantic memory is developed with pose-graph-aware reintegration of historical observations and keyframe-guided RMP, enabling task-relevant 3D semantic evidence to remain aligned under pose-graph corrections and object relocation while keeping observation history bounded.
        \item A hybrid target-localization and task-oriented navigation pipeline is developed by combining language-conditioned 3D retrieval, open-vocabulary image detection, mLLM-based verification, value-guided exploration, and manipulation-aware docking.
        \item The integrated system is evaluated on a real mobile manipulation platform in four dynamic indoor scenes, with task success, memory usage, update time, and module ablations reported against representative baselines.
    \end{itemize}

\section{Related Work}

\subsection{Open-Vocabulary Mobile Manipulation}
Open-vocabulary mobile manipulation enables robots to execute user-specified instructions in everyday environments, such as ``pick A from B and place C''~\cite{ok-robot}. Large language models (LLMs)~\cite{gpt, gemini} and vision-language models (VLMs) are commonly used to parse the instruction and decompose it into subtasks~\cite{taskplanning1}. Unlike pure navigation~\cite{navigation1}, mobile manipulation requires the robot to approach target objects, maintain sufficient visibility and reachability, and operate safely in cluttered surroundings.
Deep-reinforcement-learning approaches~\cite{maniskill-hab, amo} and imitation-learning systems~\cite{mobilealoha, beyondmimic} have shown promising manipulation skills, but their real-world generalization remains limited when object categories, layouts, or environments differ from training data. Modular frameworks~\cite{ok-robot, dovsg, dynamem} that combine language understanding, environment mapping, planning, and low-level control are therefore attractive for real robotic systems. \ourname{} follows this modular paradigm but focuses on dynamic target relocation, online memory maintenance, and active target reacquisition in previously unseen indoor environments.

\subsection{3D Semantic Scene Representation}
Mainstream semantic 3D scene representations can be broadly divided into voxel/field representations~\cite{ok-robot, clipfields} and object-level scene graphs~\cite{conceptgraphs, roboexp, dovsg}. A 3D voxel or semantic field assigns features or labels to points/voxels and therefore supports direct object localization. However, dense spatial representations can require substantial memory in large-scale environments and may not explicitly encode object relations needed by high-level planning.
In contrast, 3D scene graphs model objects as nodes and spatial or semantic relations as edges, explicitly capturing scene structure in a graphical form. This makes them particularly attractive for hierarchical navigation and large-scale navigation tasks. Nonetheless, when it comes to fusing geometry and semantics, constructing and maintaining a consistent 3D scene graph is computationally expensive. In tasks that require real-time scene reconstruction and continuous detection of environmental changes, their efficiency is typically much lower than that of 3D voxel maps.
Recent work has leveraged NeRF~\cite{rosinol2022nerfslamrealtimedensemonocular, loc-nerf} or 3D Gaussian Splatting~\cite{3dgaussiansplatting, dynamicgsgdynamic3dgaussian} for scene reconstruction. While these methods can produce high-fidelity representations, real-time constraints generally force the assumption of a static environment. Extending them to dynamic scenes usually demands substantial computational and time overheads, making them difficult to deploy for online mobile manipulation.
The relationship among representative open-vocabulary representations is important for positioning \ourname{}. ConceptGraphs~\cite{conceptgraphs} primarily builds an object-centric scene graph under a static-scene assumption. DovSG~\cite{dovsg} extends this line toward dynamic object-level scene graphs. CLIP-Fields~\cite{clipfields} is a representative semantic-field representation, while DynaMem~\cite{dynamem} extends voxel/semantic-field memory toward online mobile manipulation in changing scenes. \ourname{} adopts a voxel map as a pragmatic representation for online operation, but augments it with pose-graph-aware reintegration, stale-observation removal, and keyframe-guided pruning so that the active memory remains compact and consistent under dynamic target relocation.

\subsection{Mapping and Navigation in Dynamic Environments}
For navigation, Simultaneous Localization and Mapping (SLAM) is fundamental, as it determines the quality of geometric and semantic reconstruction and thus impacts downstream planning and manipulation. Some recent works evaluate purely in simulation under known or near-perfect robot state \cite{wu2025embodiedinstructionfollowingunknown, yin2025unigoaluniversalzeroshotgoaloriented}; others discard explicit maps and rely only on images \cite{Haviland_2022}, using large models to output end-to-end actions or short-horizon policies \cite{cheng2025navilaleggedrobotvisionlanguageaction}. Although these approaches are promising in certain regimes, the lack of an explicit, maintainable scene memory makes it difficult to guarantee consistent behavior over long horizons in complex real-world environments.
DynaMem~\cite{dynamem} brings voxel-based semantic mapping to real-world mobile manipulation by using a SLAM system to provide camera poses and a 3D voxel memory to detect scene changes and trigger updates, which in turn supports downstream replanning. However, when historical camera poses are later refined by loop closure or pose-graph optimization, previously integrated semantic observations may become misaligned with the updated global trajectory. DynaMem does not explicitly maintain and reintegrate historical semantic observations under such pose corrections. In contrast, \ourname{} associates stored observations with pose-graph keyframes and uses optimized historical poses to selectively rebuild the active spatio-semantic memory. This pose-graph-aware memory maintenance, together with keyframe-guided pruning, keeps the memory compact and temporally consistent for long-horizon target localization, reacquisition, and navigation-manipulation in dynamic environments.

\section{Method}
 This section presents the overall architecture of \ourname{} and its three main modules: (i) a multi-sensor SLAM backend with a dynamic spatio-semantic voxel memory and redundancy-aware pruning, (ii) a hybrid localization and task-oriented navigation module, and (iii) a grasping and placement stack for mobile manipulation in dynamic, previously unseen indoor environments without any pre-built map.

\begin{figure*}
    \centering
    \includegraphics[width=1\linewidth]{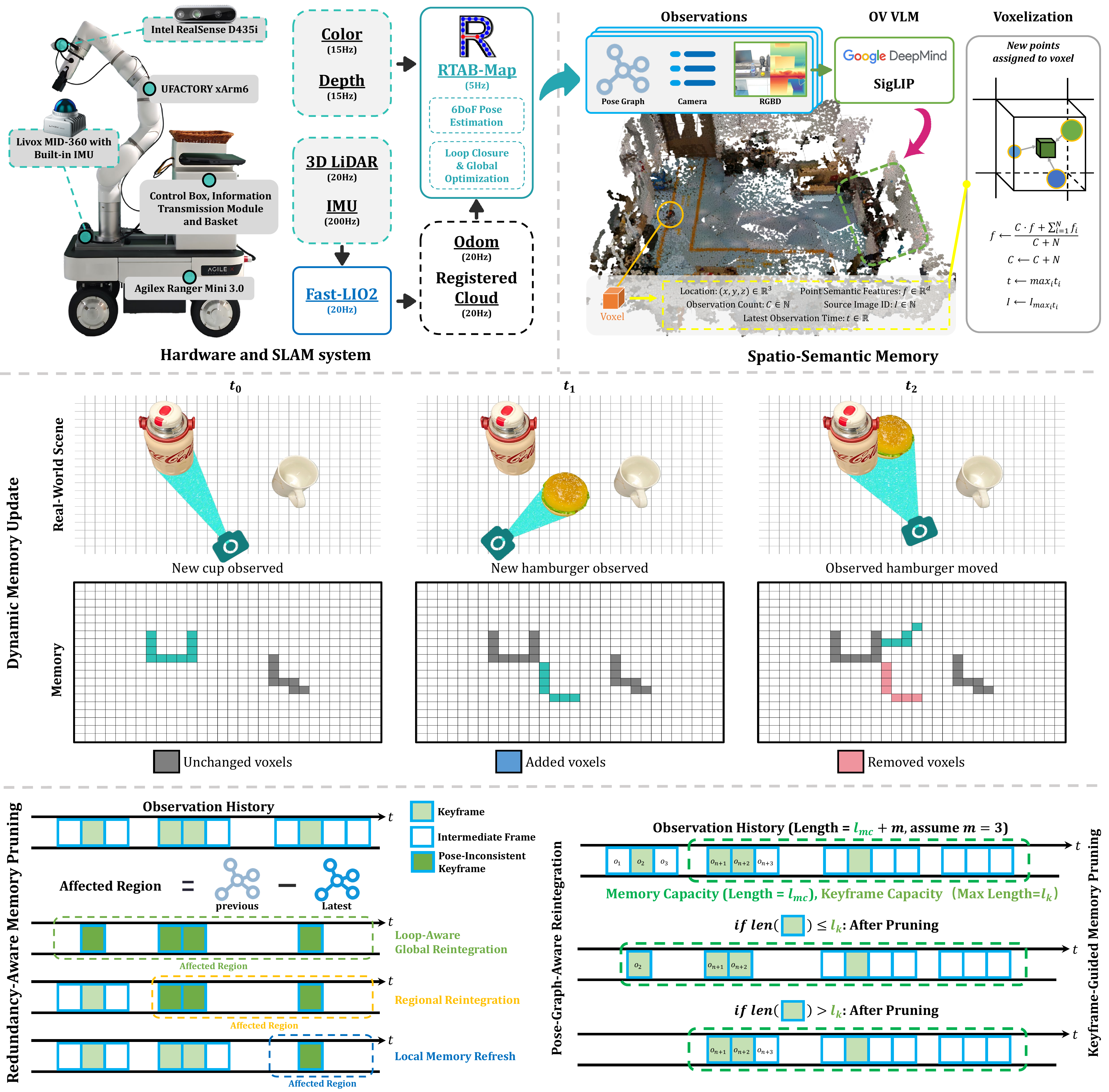}
    \vspace{-8mm}
    \caption{\textbf{Hardware system and dynamic spatio-semantic memory in \ourname{}.} The multi-sensor SLAM backend provides temporally consistent poses to build a voxel-based spatio-semantic memory, which is continuously updated with incoming observations. A Redundancy-Aware Memory Pruning mechanism performs pose-graph-aware reintegration and keyframe-guided pruning to maintain a compact and consistent map in dynamic environments.}
    \label{fig:hs_dssm}
    \vspace{-4mm}
\end{figure*}

\subsection{Multi-Sensor SLAM and Dynamic Spatio-Semantic Memory}
\label{method_subsec2}

To support long-horizon mobile manipulation in dynamic, previously unseen environments, \ourname{} uses a practical multi-sensor geometric backend and a 3D spatio-semantic memory. This subsection describes the LiDAR-inertial-visual SLAM pipeline that provides camera poses and registered geometry for downstream perception, the voxel-based spatio-semantic memory built on top of this pipeline, and the Redundancy-Aware Memory Pruning (RMP) mechanism that keeps the memory compact while preserving long-horizon consistency.

\subsubsection{\textbf{ Multi-Sensor SLAM Backend}}

 The platform uses an RGB-D camera and a 3D LiDAR with a built-in IMU to provide geometric and visual information for mapping and localization (see the top-left part of Fig.~\ref{fig:hs_dssm}). The extrinsic transforms between the camera, LiDAR, and robot base are exported from the robot CAD model, and sensor streams are aligned using software timestamps. This deployment choice avoids specialized calibration and hardware synchronization requirements; remaining practical residual errors are reduced by the downstream visual loop closure and pose-graph refinement in RTAB-Map~\cite{rtabmap}.

For multi-sensor fusion, LiDAR and IMU measurements from the MID-360 are processed by Fast-LIO2 ~\cite{fastlio2} to obtain local LiDAR-inertial odometry and de-skewed registered point clouds. These odometry estimates, together with RGB and depth images from the D435i, are provided to RTAB-Map~\cite{rtabmap}, which performs RGB-D loop closure and pose-graph optimization to produce temporally consistent camera trajectories. RTAB-Map organizes incoming frames into keyframes and intermediate frames: keyframes are inserted into the pose graph for long-term optimization, whereas intermediate frames are used locally and then discarded. The optimized camera poses and RTAB-Map keyframes are then consumed by the spatio-semantic memory and by RMP. In this work, the SLAM backend serves as a practical geometric foundation for online memory maintenance, target reacquisition, navigation, and manipulation.

\subsubsection{\textbf{Dynamic Spatio-Semantic Memory}}
 \ourname{} constructs an updatable spatio-semantic memory that maintains geometric and open-vocabulary semantic information of the environment. A sparse 3D voxel map is used as the representation. Each non-empty voxel stores its 3D location $(x, y, z)$, observation count $C$, source image ID $I$, semantic feature vector $f$ extracted from a VLM such as CLIP~\cite{clip} or SigLIP~\cite{zhai2023sigmoidlosslanguageimage}, and latest observation time $t$, following the general voxel-memory design of DynaMem~\cite{dynamem}. The dynamic update process consists of adding new observations and removing outdated voxels whose corresponding surfaces have been moved or removed.

\textbf{Adding Points.} When \ourname{} receives a new set of observations from RTAB-Map, i.e., RGB-D images, camera parameters, and pose-graph information, the depth image is back-projected into the global reference frame and a semantic feature vector is generated for each point. Pixel-level semantic features from SigLIP ~\cite{zhai2023sigmoidlosslanguageimage} are associated with the corresponding 3D points; details of mapping image features to 3D points follow prior work ~\cite{ok-robot, dovsg}. The new points are clustered and assigned to their nearest voxel grid cells. As shown in the top-right of Fig.~\ref{fig:hs_dssm}, each voxel count tracks the total number of assigned points, while the feature vector stores the weighted average of all assigned features. The observation timestamp and image ID are also updated so that the memory can identify recent evidence for future localization and pruning. If a voxel is empty before assignment, its count is initialized as $C=0$ and its feature vector $f$ is the zero vector.

\textbf{Removing Points.} When an object is moved or removed in a real-world environment, its associated voxels in \ourname{} may also need to be removed. Ray casting is used to identify obsolete voxels: if a voxel lies inside the frustum between the camera plane and the currently observed surface, that voxel should be empty in the updated scene. To reduce the impact of long-range depth noise and ensure that the map built from the current frame supports short-range navigation, pixels whose associated depth values exceed 3 \,m are ignored. A simplified 2D illustration is shown in the middle of Fig.~\ref{fig:hs_dssm}. In practice, each existing voxel is projected onto the camera plane and its distance to the camera is computed. Suppose the image has height and width $(H, W)$, and the depth image is $\mathbf{D}$. If a voxel is projected to image coordinates $(h, w)$ on the camera plane with depth $d$, then the voxel is removed when $d$ satisfies the following two conditions.
\begin{equation}
    0 \le h < H,\quad 0 \le w < W
    \label{eq:pixel_range}
\end{equation}
\begin{equation}
    d \in \bigl(0, \min(3, \mathbf{D}[h, w] + \epsilon)\bigr)
    \label{eq:depth_range}
\end{equation}

Eq.~\ref{eq:pixel_range} guarantees that the projected pixel falls inside the camera view, whereas Eq.~\ref{eq:depth_range} enforces that (a) $d > 0$ so the voxel is in front of the camera, (b) $d < 3$\,m so it is not excessively far away, and (c) $d < \mathbf{D}[h, w] + \epsilon$ so it lies between the camera and the surface observed at pixel $(h, w)$ up to a small depth tolerance $\epsilon$.

\subsubsection{\textbf{Redundancy-Aware Memory Pruning}}
\ourname{} maintains a spatio-semantic voxel memory that grows over time as new observations are integrated. Redundancy-Aware Memory Pruning (RMP) is designed as a memory-maintenance mechanism rather than only a frame-discarding strategy. It performs two coupled operations: pose-graph-aware refreshing of historical observations and keyframe-guided pruning of redundant observation history. Each stored observation keeps the pose and frame-transform information needed to reconstruct its camera pose when the SLAM pose graph is updated. RMP compares the pose used during the original integration of a keyframe with the corresponding optimized pose from the updated pose graph. When the change is sufficiently large, the affected part of the active spatio-semantic memory is refreshed by reprocessing the relevant historical observations with their optimized poses. This allows the semantic memory to follow loop-closure and pose-graph corrections, rather than accumulating stale semantic evidence at outdated positions. All observations that appear as nodes in the pose graph are treated as keyframes, and the keyframe set is denoted by $\mathcal{K}$. For each keyframe, the camera pose used during its original integration is compared with the optimized pose from the pose graph.
Let $\Delta\mathbf{t}$ and $\Delta\theta$ denote the translation and rotation differences; if $\|\Delta\mathbf{t}\|_2 > \tau_{\text{trans}}$ or $\Delta\theta > \tau_{\text{rot}}$, where $\tau_{\text{trans}}$ and $\tau_{\text{rot}}$ are translational and rotational thresholds, the keyframe is marked as pose-inconsistent and added to $\mathcal{K}_{\text{u}} \subset \mathcal{K}$, as illustrated in Fig.~\ref{fig:hs_dssm}.

\textbf{Pose-Graph-Aware Reintegration.}
Pose-graph-aware reintegration decides which part of the observation history should be rebuilt using optimized keyframe poses. Let $w_{\text{rt}}$ and $w_{\text{sm}}$ be the lengths of the real-time and short-term windows over the keyframe history, and let $\mathcal{K}_{\text{rt}}$ and $\mathcal{K}_{\text{sm}}$ be the corresponding subsets of $\mathcal{K}$. When a large fraction of keyframes is pose-inconsistent, i.e., $|\mathcal{K}_{\text{u}}| / |\mathcal{K}| > \rho_{\text{L}}$ with $\rho_{\text{L}} = 2/3$, the system regards this as a large loop closure and resets the voxel-level semantic memory, followed by reintegrating all keyframes in chronological order using their optimized poses (\textit{Loop-Aware Global Reintegration}, see the bottom-left of Fig.~\ref{fig:hs_dssm}). If instead more than a fraction $\rho_{\text{M}} \approx 1/3$ of the keyframes in the short-term window are pose-inconsistent, i.e., $|\mathcal{K}_{\text{u}} \cap \mathcal{K}_{\text{sm}}| / |\mathcal{K}_{\text{sm}}| > \rho_{\text{M}}$, and at least one pose-inconsistent keyframe lies in the real-time window ($|\mathcal{K}_{\text{u}} \cap \mathcal{K}_{\text{rt}}| > 0$), the affected observations in the last $w_{\text{sm}}$ frames are refreshed by rebuilding the corresponding part of the active memory using their optimized poses (\textit{Regional Reintegration}). Otherwise, if $|\mathcal{K}_{\text{u}} \cap \mathcal{K}_{\text{rt}}| > 0$ but the above two conditions are not met, keyframes within $\mathcal{K}_{\text{rt}}$ are refreshed using their optimized poses to perform a lightweight local refresh (\textit{Local Memory Refresh}); otherwise, the pose update is ignored for the spatio-semantic memory.

\textbf{Keyframe-Guided Memory Pruning.}
Let $l_{mc}$ denote the memory capacity (maximum history length), $m$ the current overflow length beyond this capacity, and $l_k$ the maximum number of keyframes allowed in memory, as illustrated in the bottom-right of Fig.~\ref{fig:hs_dssm}. When the history length reaches $l_{mc} + m$, the $m$ oldest observations are removed in a keyframe-aware manner: intermediate frames are discarded first, and keyframes are removed only when their count exceeds $l_k$. Here $m$ is determined online as the current excess history length rather than tuned as an additional fixed hyperparameter. To further reduce the cost of storing and decoding high-dimensional semantic features, feature vectors are retained only for keyframes and for the most recent $w_{\text{rt}}$ observations, which serve as a buffer for real-time updates, while older non-keyframe observations keep only geometric information.
 The experimental configuration uses $\tau_{\text{trans}}=0.1$\,m, $\tau_{\text{rot}}=5^\circ$, $w_{\text{sm}}=50$, $w_{\text{rt}}=10$, $l_{mc}=500$, and $l_k=100$.
In combination, pose-graph-aware reintegration and keyframe-guided pruning keep the spatio-semantic memory compact and consistent over long horizons with low computational overhead.

\begin{figure}
    \centering
    \includegraphics[width=0.9\linewidth]{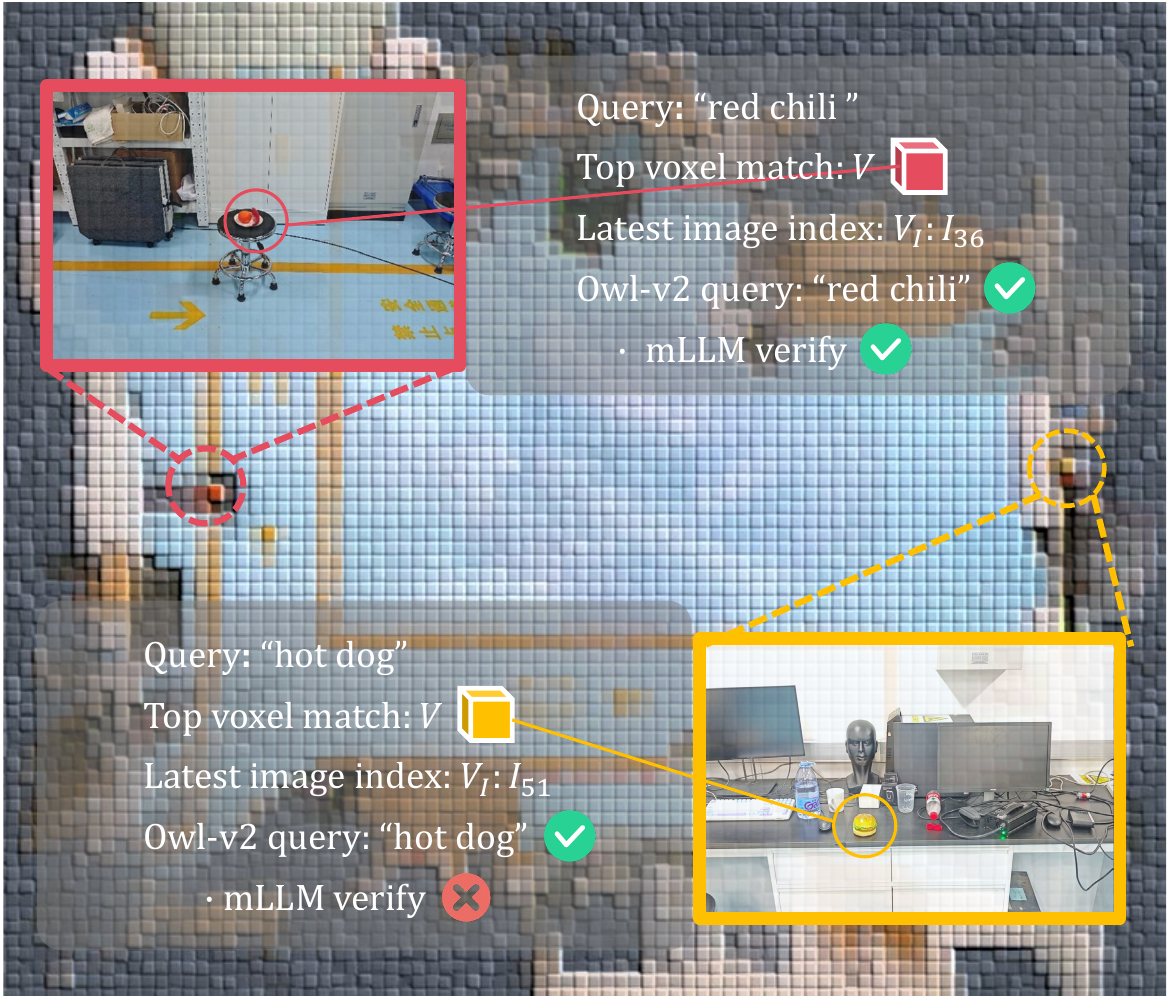}
    \vspace{-2.5mm}
    \caption{\textbf{Hybrid Localization.} The VLM feature field in the spatio-semantic memory is first queried with a language embedding to retrieve a high-confidence voxel and its associated RGB frame, which is then passed through an open-vocabulary detector to obtain object proposals. Finally, an mLLM inspects the retrieved image and either confirms or rejects the presence of the queried object, yielding a verified 3D target pose.}
    \label{fig:hl}
    \vspace{-4mm}
\end{figure}

\subsection{Hybrid Localization and Task-Oriented Navigation}
\label{method_subsec3}
 This subsection describes how \ourname{} uses dynamic spatio-semantic memory for hybrid localization, task-oriented navigation, and target reacquisition.

\subsubsection{Hybrid Localization}
\label{lab:hl}
 The hybrid localizer builds on VLMs such as CLIP~\cite{clip} and SigLIP~\cite{zhai2023sigmoidlosslanguageimage}, which embed images and text into a shared latent space for coarse memory retrieval, followed by image-space detection and high-level semantic verification.
As described in Sec.~\ref{method_subsec2}, RGB-D observations are converted into per-point visual features and integrated into the voxelized memory. Given a language query, the text is encoded by the VLM text encoder, and coarse localization identifies the voxel whose stored feature has the highest inner product with this embedding.
Conditioned on the selected voxel, the most recent associated RGB image is retrieved from the spatio-semantic memory and passed to an open-vocabulary object detector (e.g., OWL-v2~\cite{owl-v2}) to detect instances of the queried category in image space. The 3D position of the candidate object is then obtained by back-projecting the center pixel of the detected bounding box into the world frame.

 A purely similarity-threshold-based approach is brittle with semantically or visually similar objects; for example, a ``hot dog'' query may still retrieve a ``hamburger'' as a confident match.
Multimodal large language models (mLLMs) ~\cite{gpt} can better resolve such distinctions, so an mLLM-based semantic verification stage is introduced. As shown in Fig.~\ref{fig:hl}, the pipeline has three stages: (i) memory-based coarse localization, which retrieves the highest-confidence voxel and its most recent image; (ii) open-vocabulary detection on this image to generate 3D object hypotheses; and (iii) mLLM-based re-ranking and verification to reject visually inconsistent detections. This design enables \ourname{} to localize the \emph{actual} target rather than the closest semantic surrogate.

 The implementation uses SigLIP-so400m for language-conditioned feature extraction and OWL-V2 Large ensemble for open-vocabulary detection. The object and grasp feature-matching thresholds are both set to 0.05, and the OWL-V2 detector uses a confidence threshold of 0.15. Qwen-VL-Max~\cite{qwen25vl} is used as the mLLM verifier, which returns a short caption and a binary \texttt{Answer: True/False} decision for each candidate target. The decision is then parsed to accept or reject the candidate target.

If no candidate survives the verification stage, the query is treated as not grounded in the current memory, the object is marked as ``object not found'', and control is passed to the exploration policy described next. Meanwhile, RMP refreshes pose-inconsistent historical observations and prunes redundant history, so retrieval and verification operate over a bounded and temporally consistent set of voxels and images, keeping the hybrid localizer efficient even in long-horizon tasks.

\begin{figure}
    \centering
    \includegraphics[width=\linewidth]{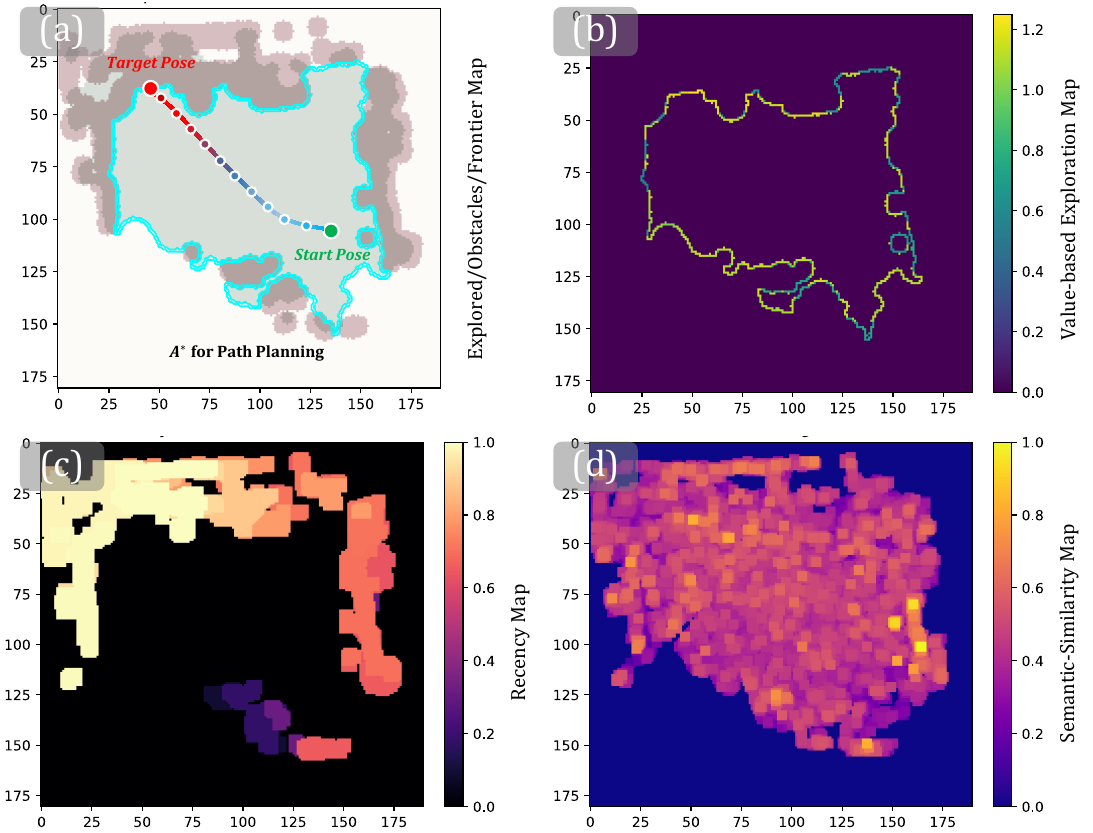}
    \vspace{-7mm}
    \caption{\textbf{Task-oriented Navigation and Exploration.} (a) A* path planning on the explored/obstacle/frontier map from the current start pose to the task-specific target pose. (b) Exploration score map obtained by combining the recency and semantic-similarity values; high-score frontier cells are selected as exploration goals. (c) Recency map $V_T$ that assigns a higher value to regions that have not been observed for a long time. (d) Semantic-similarity map $V_S$ computed from VLM features, highlighting regions whose appearance is most relevant to the language query.}
    \label{fig:ton}
    \vspace{-15pt}
\end{figure}

\subsubsection{Task-Oriented Navigation and Exploration}
To realize task-oriented navigation, \ourname{} combines a grid-based A*~\cite{hart1968formal} planner with the dynamic spatio-semantic memory. Depending on whether the target is localized, the system tracks a known target pose or explores with a value-based map and distance-aware heuristics. Near the object, a \emph{docking point} is introduced for manipulation-ready base poses, and paths are executed in a receding-horizon fashion to remain robust to scene changes.

\textbf{Obstacle Map Construction.}
A 2D obstacle map is first constructed from the spatio-semantic memory. A sparse 3D point cloud is normalized with respect to the sensor height and the ground plane is translated to $z=0$. To account for noise in depth measurements, a ground threshold is used (0.2\,m in the experiments): voxels above this height are marked as obstacles, while voxels below are projected onto a 2D grid as navigable cells. Grid cells that are neither labeled as obstacles nor as navigable are treated as explorable frontier cells, from which the planner can expand the explored region.

\textbf{Exploration and Goal Selection.}
Given the robot pose and a target pose, the planner treats the target pose as the navigation goal and runs A* on the obstacle map when Hybrid Localization confidently localizes the target (Fig.~\ref{fig:ton}(a)). If Hybrid Localization returns ``object not found'', \ourname{} switches to exploration mode and selects frontier cells as intermediate goals. Rather than exploring uniformly, this frontier-based exploration is biased by a \textit{value-based exploration map} (Fig.~\ref{fig:ton}(b)) toward promising regions.
This map is decomposed into two terms. The first term is a \textit{recency map} (Fig.~\ref{fig:ton}(c)) that prioritizes locations that have not been observed for a long time. Let the current time be $T$ and the last-seen time of voxel $(x,y,z)$ be $t_{x,y,z}$. The recency map $V_T$ is defined as
\begin{align}
T^*[x,y] &= \max_{z} \bigl(T - t_{x,y,z}\bigr),\\
V_T[x,y] &= \sigma\!\left(\beta_T\bigl(T^*[x,y]-\mu_T\bigr)\right),
\end{align}
where $\beta_T$ and $\mu_T$ are hyperparameters, and $\sigma$ denotes the sigmoid function.
The second term is a \textit{semantic-similarity map} (Fig.~\ref{fig:ton}(d)) that exploits the VLM features stored in the spatio-semantic memory. Let $f_{x,y,z}$ be the VLM feature at voxel $(x,y,z)$ and $f_q$ be the VLM feature of the language query. The semantic-similarity map $V_S$ is defined as
\begin{align}
S^*[x,y] &= \max_{z}\bigl(f_q \cdot f_{x,y,z}\bigr),\\
V_S[x,y] &= \sigma\!\left(\beta_S\bigl(S^*[x,y]-\mu_S\bigr)\right),
\end{align}
where $\beta_S$ and $\mu_S$ are also hyperparameters. A linear combination of $V_T$ and $V_S$ defines the exploration score map in Fig.~\ref{fig:ton}(b), balancing temporal novelty and semantic relevance.

During exploration, a distance-dependent heuristic is further imposed to reduce local oscillations when sampling exploration goals from this score map. Let $p_t \in \mathbb{R}^2$ denote the robot base position at time $t$, and let $\mathcal{G}_t$ be the set of frontier cells sorted by their exploration score. Frontier cells that lie outside a small neighborhood of the base are first identified as
\begin{equation}
    \mathcal{F}_t
    =
    \left\{
        g \in \mathcal{G}_t \;\middle|\;
        \|g - p_t\|_2 \ge d_{e}
    \right\},
\end{equation}
with $d_{e} = 0.85\,\text{m}$ in the implementation. In practice, the planner samples the next exploration goal from
\begin{equation}
    \mathcal{G}_t^{\text{explore}} =
    \begin{cases}
        \mathcal{F}_t, & \text{if } \mathcal{F}_t \neq \emptyset,\\[1mm]
        \mathcal{G}_t, & \text{otherwise},
    \end{cases}
    \label{eq:explore_goals}
\end{equation}
 Thus, the planner prefers high-score frontier cells at least $d_e$ away from the current base when available, and otherwise falls back to all frontier cells. This reduces local goal oscillations near the robot.

\textbf{Docking and Receding-horizon Execution.}
Once the target object has been localized, navigation near the goal is driven by manipulation requirements rather than pure coverage. Let $o_t \in \mathbb{R}^2$ denote the 2D position of the queried object at time $t$. A manipulation range $d_{m}$, also set to $0.85\,\text{m}$, defines a \emph{docking point}: a base pose that places the robot within reach of the object while keeping a collision-free corridor between the base and the object. Formally, let $\mathcal{D}_t$ be a set of navigable cells in a neighborhood of $o_t$; feasible docking candidates are
\begin{equation}
    \mathcal{D}_t^{\text{dock}}
    =
    \left\{
        g \in \mathcal{D}_t \;\middle|\;
        \|g - o_t\|_2 \le d_{m},\
        \text{localFree}(g, o_t)
    \right\},
    \label{eq:docking_goals}
\end{equation}
where $\text{localFree}(g, o_t)$ is a lightweight local collision check that rejects base goals whose corridor to the object is occupied in the obstacle map. In the implementation, $g_t^{\text{dock}}$ is selected from $\mathcal{D}_t^{\text{dock}}$ as the feasible goal with the lowest path cost from $p_t$, encouraging sufficient clearance and line of sight for manipulation.

 To handle scene changes, navigation is executed in a semi-closed-loop, receding-horizon fashion. Instead of executing the full A* path, the robot follows only the first 16 waypoints (about $1\text{-}1.5$\,m), then re-senses, updates the map, and replans. Once $\|p_t - o_t\|_2 \le d_m$, this truncation is disabled to avoid unnecessary detours near the goal. The cycle repeats until the robot is close enough to the target and no obstacles block manipulation; the base is then aligned to face the object and the camera centers the target in the image.

 The receding-horizon strategy does not discard useful target history. When a target has been localized, \ourname{} caches the corresponding navigation goal and plans toward it while continuing to verify the scene. In the released configuration, once the robot is within 2.0\,m of the cached target, it switches from a full look-around sweep to focused verification toward that goal. If the new detection differs from the cached goal by more than 0.50\,m, the old candidate is treated as stale and the system replans toward the updated detection. This lets the robot exploit long-range target observations while still rechecking near the target to handle relocation.

\begin{figure}[t]
    \centering
    \includegraphics[width=\linewidth]{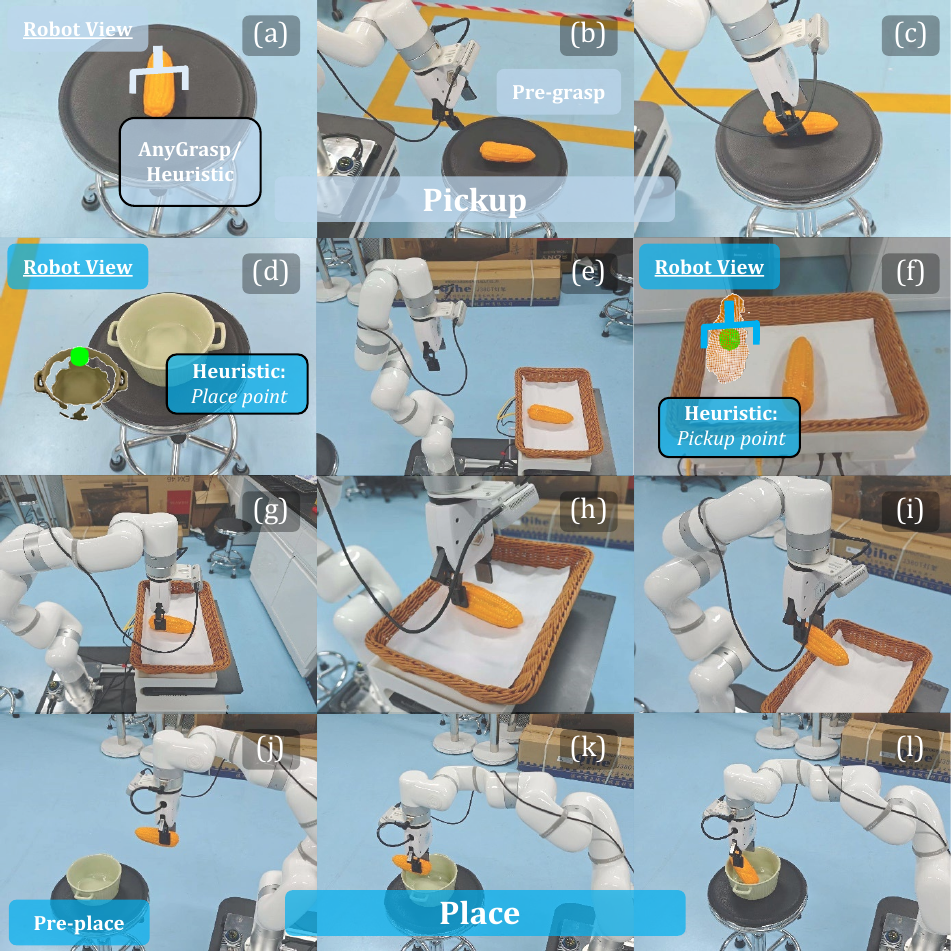}
    \vspace{-7mm}
    \caption{\textbf{Grasping and placement in \ourname{}.}
     (a)--(c) AnyGrasp / heuristic grasp proposal, pre-grasp pose, and executed pickup.
     (d)--(f) Heuristic selection of the placement point in the bowl and the pickup point in the basket.
     (g)--(i) Two-stage approach and grasp execution inside the basket.
     (j)--(l) Pre-place pose, descent, and final placement of the object in the bowl.}
    \label{fig:pick_and_place}
    \vspace{-3mm}
\end{figure}

\subsection{Grasping and Placement Strategy}
\label{method_subsec4}
Manipulation requires the base to approach the target with sufficient visibility, reachability, and collision-free arm workspace. Once the robot reaches a suitable docking point and the target is visible to the wrist-mounted RGB-D camera, \ourname{} triggers the grasping or placement routine in Fig.~\ref{fig:pick_and_place}.

\subsubsection{Grasping Strategy}
Despite recent progress in Vision Language Action (VLA) Models ~\cite{wang2025unifiedvisionlanguageactionmodel, liao2025genieenvisionerunifiedworld}, their robustness and open-vocabulary grasping capability in real-world settings remain limited. \ourname{} therefore adopts a modular grasping stack with three components: (i) AnyGrasp-based grasp proposal generation, (ii) a heuristic fallback for top-down grasps, and (iii) a two-stage execution policy.

\textbf{AnyGrasp.}
Given a language-specified target, once the robot has navigated to a docking point (Sec.~\ref{method_subsec3}), the wrist-mounted RGB-D camera is oriented toward the object and captures a local RGB-D observation. Grounding DINO~\cite{groundingdino} and Segment Anything-2~\cite{sam2} segment the queried object, and the masked RGB-D pixels are back-projected into a 3D point cloud in the robot base frame. AnyGrasp~\cite{fang2023anygrasprobustefficientgrasp} then generates collision-free grasp candidates, which are further filtered using the segmentation mask, retaining only those whose contact regions lie on the target object. The remaining candidate with the highest confidence is executed, yielding target-specific grasps rather than grasps on nearby distractors.

\textbf{Heuristic grasp.}
In real-world experiments, AnyGrasp can fail on specular or noisy surfaces where depth is unreliable. A top-down fallback is then invoked, choosing a yaw angle that closes the gripper along the shorter footprint axis of the segmented object point cloud.
 The segmented object is projected into the horizontal plane, and the median $(x,y)$ is subtracted to obtain centered points $p_i \in \mathbb{R}^2$. From these points, the covariance matrix and its two orthogonal eigenvectors $e_1,e_2$ are computed. For each axis $e_k$, the extent is measured as
\begin{equation}
    \ell_k =
    \max_i (p_i \cdot e_k)
    -
    \min_i (p_i \cdot e_k), \quad k \in \{1,2\}.
\end{equation}
The gripper closing direction is set to the shorter axis,
\begin{equation}
    c =
    \begin{cases}
        e_1, & \ell_1 \le \ell_2,\\[0.3ex]
        e_2, & \text{otherwise},
    \end{cases}
    \qquad
    \theta_{\text{yaw}} = \operatorname{atan2}(c_y, c_x),
\end{equation}
where $(c_x,c_y)$ are the components of $c$ and $\theta_{\text{yaw}}$ is wrapped to $(-\pi,\pi)$. For elongated objects, this aligns the closing direction with the shorter axis of the footprint; for nearly isotropic shapes, either axis is acceptable. In practice, this provides a simple shape-adaptive yaw estimate whenever learned grasps are unavailable.

\textbf{Two-stage execution.}
Once a grasp pose has been selected, it is executed using a short, staged approach to ensure safe and stable contact. Let $\vec{p}$ denote the grasp point in the base frame and $\vec{a}$ a unit approach direction for the grasp. The end-effector follows the trajectory
\begin{equation}
    \big\langle\,
        \vec{p} - 0.25\,\vec{a},\;
        \vec{p}
    \,\big\rangle,
\end{equation}
moving from a pre-grasp waypoint to the final grasp point, as shown in Fig.~\ref{fig:pick_and_place}(b),(c). This keeps the arm collision-free until the final approach and reduces object disturbance. The gripper then closes in 5\,mm increments while motion is monitored. When additional commanded closing produces little motion before the mechanical limit, the object is regarded as grasped. The robot then lifts it and places it into the rear storage basket.

\subsubsection{Placement Strategy}
Placement starts after the object has been stored in the rear basket and the robot is commanded to a target region, such as a bowl (Fig.~\ref{fig:pick_and_place}(d)). The procedure estimates a safe placement pose inside the target container, re-grasps the object from the basket (Fig.~\ref{fig:pick_and_place}(e)--(i)), and executes a two-stage placement motion (Fig.~\ref{fig:pick_and_place}(j)--(l)).

The robot first navigates to a docking point facing the target container and reuses the same open-vocabulary perception stack as in grasping. Grounding DINO and Segment Anything~2 segment the container, and the masked pixels are back-projected with depth to form a point cloud $\mathcal{P}^{\text{place}} = \{(x,y,z)\}$ in the robot base frame. The median horizontal coordinates $(x_m, y_m)$ provide a robust estimate of the region center. Around this center, points within a small window ($|x - x_m| < 0.1$\,m, $|y - y_m| < 0.1$\,m) are considered, and the safe placement height is defined as
\begin{equation}
    z_{\max}
    =
    0.1
    +
    \max\bigl\{
        z \;\big|\;
        |x - x_m| < 0.1,\,
        |y - y_m| < 0.1
    \bigr\},
\end{equation}
where the constant $0.1$\,m is a vertical safety margin above the local support surface. This yields a top-down placement pose at $(x_m, y_m, z_{\max})$, shown in Fig.~\ref{fig:pick_and_place}(d).

After fixing the placement pose, the arm is redirected toward the rear basket, and the same heuristic top-down grasping procedure as in the pickup phase is reused to grasp the object from the basket, as illustrated in Fig.~\ref{fig:pick_and_place}(f). Finally, with the object firmly held, the arm moves from the basket to the pre-computed placement pose using the same two-stage pattern as in grasping. It first goes to a pre-place waypoint above $(x_m, y_m, z_{\max})$, then follows a straight-line trajectory to the final pose before opening the gripper (Fig.~\ref{fig:pick_and_place}(j)--(l)), which minimizes lateral motion and reduces collisions with the container or surrounding clutter.

\section{Experiments}
\label{sec:experiments}
\ourname{} is evaluated on a real mobile manipulation platform in four dynamic indoor laboratory scenes. The evaluation focuses on three aspects: target navigation and exploration, object pickup and placement, and spatio-semantic memory efficiency. The experiments include (i) closed-loop mobile manipulation under dynamic target relocation, (ii) memory consumption and update efficiency under identical RGB-D and pose streams, and (iii) ablation studies for grasping and hybrid localization modules.

\begin{figure}[t]
    \centering
    \includegraphics[width=\linewidth]{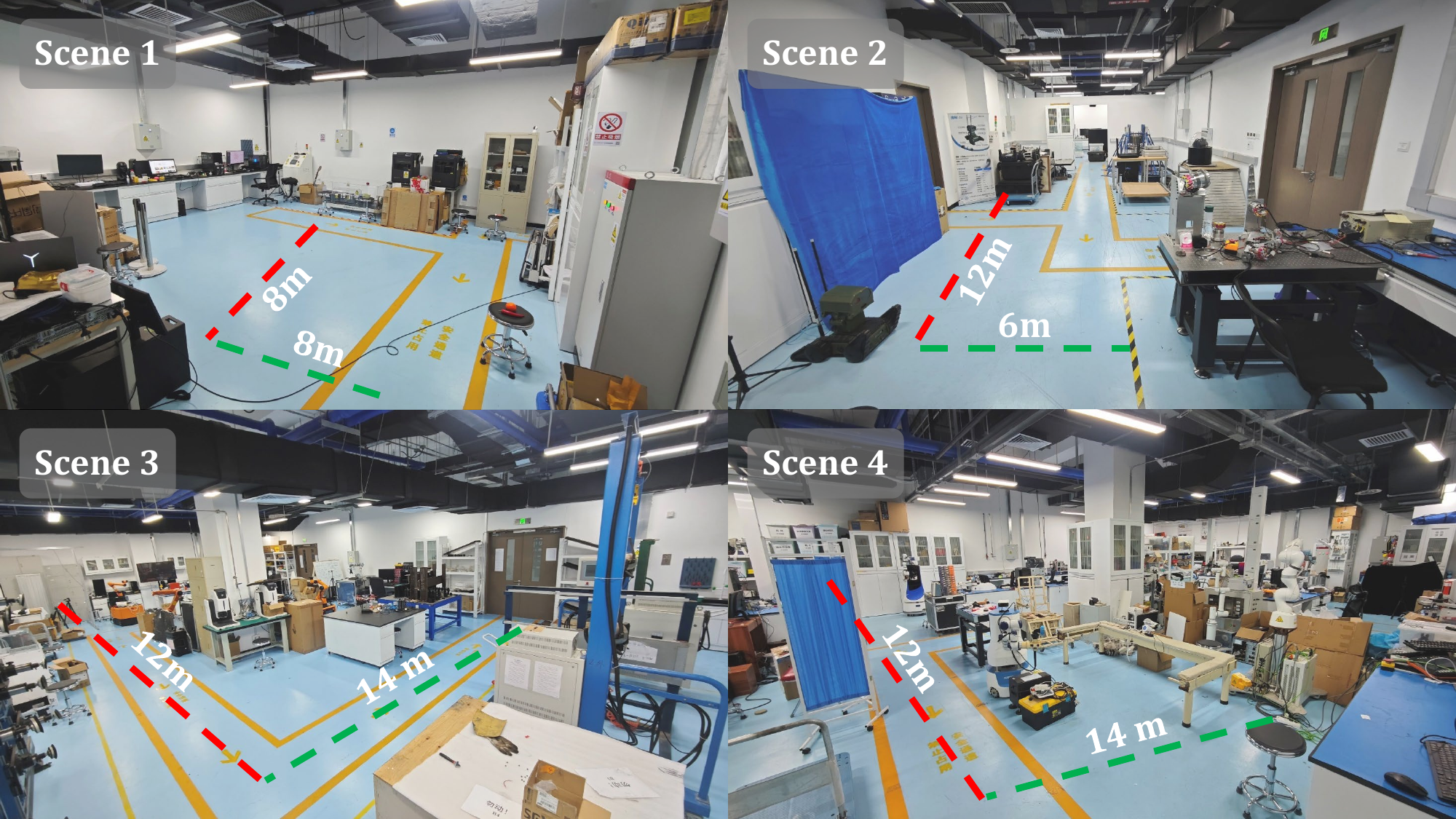}
    \vspace{-7mm}
    \caption{\textbf{Real-world laboratory scenes used for evaluation.}
    S1--S4 are dynamic indoor rooms of size $8\!\times\!8$, $6\!\times\!12$, $12\!\times\!14$, and $12\!\times\!14$\,m, respectively, with obstacle density and corridor narrowness increasing from S1 to S4.}
    \label{fig:exp_scenes}
    \vspace{-4mm}
\end{figure}

\subsection{Hardware Platform and Real-world Environments}
The mobile manipulation platform (Fig.~\ref{fig:hs_dssm}) consists of an AgileX Ranger Mini 3.0 base, a UFACTORY xArm6 manipulator, a wrist-mounted Intel RealSense D435i RGB-D camera, and a Livox MID-360 LiDAR with a built-in IMU. The robot-side system runs on a laptop with Ubuntu 22.04, ROS~2 Humble, and 64\,GB system RAM; compute-intensive perception and grasping services run on a server equipped with an NVIDIA RTX 4090 GPU with 24\,GB VRAM.
For the closed-loop mobile-manipulation comparison, DREAM and DynaMem use the same low-level control stack and the same LiDAR-inertial-visual SLAM backend.
Experiments are conducted in four real-world laboratory scenes (S1--S4; $64$, $72$, $168$, and $168\,\mathrm{m}^2$; Fig.~\ref{fig:exp_scenes}) with different obstacle densities and corridor widths. Everyday objects are randomly placed as manipulation targets. During execution, the target object is manually relocated after initial localization to evaluate memory update, target reacquisition, replanning, and task completion under scene changes.
The runtime is decoupled: RTAB-Map observations are processed at 5\,Hz, low-level control runs at 20\,Hz, servo observations at about 10\,Hz, RGB-D input at 15\,Hz, and Fast-LIO2 odometry at 20\,Hz. Semantic-memory updates and RMP maintenance run asynchronously and do not block odometry, control, or command execution.

\subsection{Mobile Manipulation in Dynamic Environments}
\label{sec:dynamic_mm}

In this task, given a natural-language instruction (e.g., ``put the orange into the green bowl''), the robot must operate in an unknown and dynamically changing environment to complete the full pipeline from task-oriented navigation and exploration to pickup and placement. Each complete long-horizon task is executed as a cascade of Navigation, Pickup, second Navigation, and Place; if one stage fails, subsequent stages in that trial are not executed. The following four evaluation metrics are used: \textbf{Navigation Success Rate:} The robot must actively explore and navigate from the initial pose to the vicinity of the target. Navigation is regarded as successful if, when the robot stops, the Euclidean distance to the target is less than 1\,m and the target is fully visible in the wrist camera image, even when the target has been moved one or more times by a human during execution. \textbf{Pickup Success Rate:} Conditioned on successful navigation, a stage attempt is counted as successful if the robot stably grasps the target and transports it into the rear storage basket. \textbf{Place Success Rate:} Conditioned on the preceding stages reaching placement, a stage attempt is counted as successful if the robot places the target stably inside or on the specified container. \textbf{Long-term Success Rate:} Each trial consists of Navigation, Pickup, a second Navigation, and Place. Long-term success requires all four stages to succeed. Since each long-horizon trial contains two navigation stages, Navigation Success Rate is computed over all first- and second-navigation attempts that are reached. Navigation, Pickup, and Place percentages are computed over the valid attempts that reach the corresponding stage, whereas Long-term success is computed over complete task trials. Raw counts are reported with percentages in Tab.~\ref{tab:dynamic_mm_results} to make these denominators explicit.

\begin{table}[t]
\centering
\caption{\textbf{Success rates (\%) on dynamic mobile manipulation tasks.}
\ourname{} is compared with DynaMem across four real-world scenes. Subtask rates use stage-conditioned denominators, Long-term success requires all stages to succeed, and raw counts are shown in parentheses.}
\label{tab:dynamic_mm_results}
\resizebox{\linewidth}{!}{
\begin{tabular}{llcccc}
\toprule
Task & Method & S1 & S2 & S3 & S4 \\
\midrule
\multirow{2}{*}{Navigation}
 & DynaMem & 85.7 {\scriptsize (30/35)} & 82.4 {\scriptsize (28/34)} & 72.7 {\scriptsize (24/33)} & 68.8 {\scriptsize (22/32)} \\
 & \textbf{\ourname{}} & \textbf{89.2} {\scriptsize (33/37)} & \textbf{86.1} {\scriptsize (31/36)} & \textbf{83.8} {\scriptsize (31/37)} & \textbf{80.0} {\scriptsize (28/35)} \\
\midrule
\multirow{2}{*}{Pickup}
 & DynaMem & 88.2 {\scriptsize (15/17)} & 87.5 {\scriptsize (14/16)} & 92.9 {\scriptsize (13/14)} & 92.3 {\scriptsize (12/13)} \\
 & \textbf{\ourname{}} & \textbf{94.4} {\scriptsize (17/18)} & \textbf{94.1} {\scriptsize (16/17)} & \textbf{94.4} {\scriptsize (17/18)} & \textbf{93.8} {\scriptsize (15/16)} \\
\midrule
\multirow{2}{*}{Place}
 & DynaMem & 92.3 {\scriptsize (12/13)} & 91.7 {\scriptsize (11/12)} & 80.0 {\scriptsize (8/10)} & 88.9 {\scriptsize (8/9)} \\
 & \textbf{\ourname{}} & \textbf{93.3} {\scriptsize (14/15)} & \textbf{92.9} {\scriptsize (13/14)} & \textbf{92.3} {\scriptsize (12/13)} & \textbf{91.7} {\scriptsize (11/12)} \\
\midrule
\multirow{2}{*}{Long-term}
 & DynaMem & 60.0 {\scriptsize (12/20)} & 55.0 {\scriptsize (11/20)} & 40.0 {\scriptsize (8/20)} & 40.0 {\scriptsize (8/20)} \\
 & \textbf{\ourname{}} & \textbf{70.0} {\scriptsize (14/20)} & \textbf{65.0} {\scriptsize (13/20)} & \textbf{60.0} {\scriptsize (12/20)} & \textbf{55.0} {\scriptsize (11/20)} \\
\bottomrule
\end{tabular}}
\vspace{-10pt}
\end{table}

\textbf{Results.}
The quantitative results are summarized in Tab.~\ref{tab:dynamic_mm_results}. Across all four scenes, \ourname{} achieves higher Navigation, Pickup, Place, and Long-term success rates than DynaMem. Navigation increases from $68.8\%\text{-}85.7\%$ with DynaMem to $80.0\%\text{-}89.2\%$ with \ourname{}, with the largest gains observed in the cluttered S3/S4 scenes. Pickup success improves from $87.5\%\text{-}92.9\%$ to $93.8\%\text{-}94.4\%$, and Place success improves from $80.0\%\text{-}92.3\%$ to $91.7\%\text{-}93.3\%$. For the complete cascade, \ourname{} reaches $55\%\text{-}70\%$ Long-term success, compared with $40\%\text{-}60\%$ for DynaMem. Aggregating the Long-term trials over all scenes, \ourname{} achieves $50/80=62.5\%$ success, compared with $39/80=48.8\%$ for DynaMem. The corresponding Wilson 95\% confidence intervals are approximately $51.5\%$--$72.3\%$ and $38.1\%$--$59.5\%$, respectively, reflecting the modest sample size of physical real-robot trials. These repeated gains across scenes suggest that online memory update, target reacquisition, and staged manipulation improve robustness when the target is manually relocated during execution.

\subsection{Memory Consumption and Update Efficiency}
To fairly compare the memory consumption and update efficiency of different scene representations under identical conditions, a log-and-replay protocol is adopted. In each scene, the robot is first teleoperated along a predefined exploration trajectory while a unified SLAM system records a continuous stream of RGB-D observations and corresponding camera/base poses. During logging, some target objects are moved to introduce human-induced scene dynamics. No scene representation is built during this data-collection phase.
The same RGB-D and pose sequence is then replayed from an empty map into DovSG~\cite{dovsg}, DynaMem~\cite{dynamem}, and \ourname{}. Each method incrementally reconstructs and updates its own scene representation frame by frame. For each scene, the final memory footprint and the average per-frame update time are recorded over the entire sequence. All methods use a spatial resolution of 0.05\,m. This log-and-replay protocol removes trajectory differences and isolates memory-update behavior; identical RGB-D observations and SLAM poses are provided to all memory baselines, so the reported differences reflect memory-maintenance and update behavior rather than different pose inputs. In live operation, \ourname{} uses the same memory-update logic asynchronously as described above.

\textbf{Results.}
Tab.~\ref{tab:memory_time} reports the memory footprint and average per-frame update time for DovSG, DynaMem, and \ourname{} across all four scenes.
As the number of processed frames increases, the voxel count and memory usage of DynaMem grow rapidly, reaching up to $10.43$\,GB in S4.
Its per-frame update time also increases to more than $2$\,s in the larger scenes.
DovSG maintains a more compact memory footprint (below $1$\,GB in all scenes) thanks to its node-level semantic representation, but its per-frame update time exceeds $110$\,s, dominated by repeated calls to heavy vision-language models and graph maintenance.
This makes it impractical for real-time online operation on the robotic platform.
In contrast, \ourname{} keeps both memory and update time nearly flat as the task progresses.
Across all four scenes, \ourname{} uses only $0.37$--$0.63$\,GB of memory, which is roughly $3$-$15\times$ smaller than DynaMem at the same map resolution, and achieves average per-frame update times of $0.43$--$0.53$\,s, which are about $2$-$4\times$ faster than DynaMem, even as loop closures and dynamic scene changes accumulate over long trajectories.
These results show that the proposed Redundancy-Aware Memory Pruning allows \ourname{} to maintain a compact and consistent spatio-semantic representation while keeping memory and computation bounded during long-term operation.

\begin{table}[t]
\centering
\caption{\textbf{Memory usage and per-frame update time across scenes.}
Memory is measured in GB, time in seconds (s). Lower is better.}
\label{tab:memory_time}
\resizebox{\linewidth}{!}{
\begin{tabular}{lcccccccc}
\toprule
& \multicolumn{4}{c}{Memory (GB)$\downarrow$} & \multicolumn{4}{c}{Time (s)$\downarrow$} \\
\cmidrule(lr){2-5}\cmidrule(lr){6-9}
Method & S1 & S2 & S3 & S4 & S1 & S2 & S3 & S4 \\
\midrule
DovSG
  & \textbf{0.23} & \textbf{0.35} & 0.83 & 0.89
  & 113 & 115 & 124 & 128 \\
DynaMem
  & 1.32 & 2.45 & 8.96 & 10.43
  & 0.82 & 1.11 & 2.21 & 2.38 \\
\textbf{\ourname{}}
  & 0.37 & 0.46 & \textbf{0.61} & \textbf{0.63}
  & \textbf{0.43} & \textbf{0.49} & \textbf{0.53} & \textbf{0.52} \\
\bottomrule
\end{tabular}}
\vspace{-10pt}
\end{table}

\subsection{Ablation Studies}

To analyze the contribution of key components in \ourname{}, ablation studies are conducted on the grasping and hybrid localization modules.

\textbf{(1) Grasping module.}
The following configurations are compared: \textit{(i) Full configuration}: AnyGrasp for grasp candidate generation + heuristic for grasp pose generation + two-stage trajectory execution; \textit{(ii) Without two-stage execution}: AnyGrasp for grasp candidate generation and the heuristic for grasp pose generation are retained, but only a single-stage grasp trajectory is executed; \textit{(iii) AnyGrasp only}: directly execute the grasp candidates output by AnyGrasp without heuristic or two-stage execution; \textit{(iv) Heuristic only}: do not use AnyGrasp; grasp poses are generated purely by the heuristic and executed directly.
These variants are evaluated in a tabletop setting with a fixed-base manipulator. Ten common household objects are used to test the grasping-related modules. During evaluation, each object is placed in different poses and is attempted ten times, resulting in 100 grasp trials per configuration.

\textbf{Results.}
Tab.~\ref{tab:ablation} (top) summarizes the Pickup success rates of different grasping configurations in a tabletop setting.
Using only the heuristic top-down grasp yields a $67\%$ success rate, while using only AnyGrasp improves this to $79\%$.
Combining AnyGrasp with the heuristic refinement (without two-stage execution) further increases the success rate to $88\%$, and the full configuration with two-stage execution reaches $94\%$.
These results indicate that (i) learned grasp proposals are important for difficult objects, (ii) the heuristic refinement helps when AnyGrasp fails or returns suboptimal poses, and (iii) the two-stage execution reduces collisions and stabilizes contact, leading to the best overall performance.

\begin{table}[t]
\centering
\caption{\textbf{Ablation study of Pickup and hybrid localization modules.}
Success rates (\%) averaged over all trials.}
\label{tab:ablation}
\resizebox{\linewidth}{!}{
\begin{tabular}{llc}
\toprule
Task & Method / Variant & Success $\uparrow$ \\
\midrule
\multirow{4}{*}{Pickup}
 & Heuristic only & 67 \\
 & AnyGrasp only & 79 \\
 & AnyGrasp + Heuristic (no two-stage) & 88 \\
 & AnyGrasp + Heuristic + two-stage (\ourname{}) & \textbf{94} \\
\midrule
\multirow{3}{*}{Query object}
 & VLM (CLIP) + OWL-V2 & 75 \\
 & VLM (SigLIP) + OWL-V2 & 77 \\
 & Full hybrid localization (\ourname{}) & \textbf{83} \\
\bottomrule
\end{tabular}}
\vspace{-10pt}
\end{table}

\textbf{(2) Hybrid localization module.}
For the hybrid localization mechanism, the following variants are evaluated: \textit{(i) VLM feature field + OWL-V2 (SigLIP)}: use a VLM feature field built from SigLIP features combined with OWL-V2 detection, and determine the target position via nearest-neighbor retrieval in the feature space; \textit{(ii) VLM feature field + OWL-V2 (CLIP)}: replace SigLIP with CLIP features while keeping the rest of the pipeline unchanged; \textit{(iii) Full hybrid localization}: build on VLM feature retrieval and OWL-V2 detection, and additionally introduce an mLLM-based semantic verification step to filter and confirm candidate targets.
The same log-and-replay protocol as the memory experiment is used to collect RGB-D and pose sequences. Before data collection, objects in each scene are arranged to satisfy two conditions: (i) the scene contains object categories that are truly present and can be used as query targets; and (ii) semantically or visually similar distractor objects (e.g., peppers of different colors from the same category, or a hamburger placed near a hot dog) are deliberately included to assess robustness in the presence of distractors. The \ourname{} pipeline then builds a scene representation from the recorded RGB-D and pose sequence. For each scene, query targets are sampled only from object categories that actually exist in that reconstructed scene. For each variant, 25 query episodes are run per scene, resulting in 100 query episodes in total across the four scenes.

\textbf{Results.}
Tab.~\ref{tab:ablation} (bottom) reports the success rates for localizing queried objects in the reconstructed scenes.
Using a CLIP-based feature field with OWL-V2 detection achieves a $75\%$ success rate.
Replacing CLIP with SigLIP provides a modest improvement to $77\%$, suggesting slightly better open-vocabulary retrieval.
The full hybrid localization of \ourname{}, which adds an mLLM-based semantic verification stage on top of the SigLIP feature field and OWL-V2 detections, achieves an $83\%$ success rate.
This shows that semantic verification effectively rejects spurious nearest-neighbor matches, especially with semantically or visually similar distractors.
Together with the feature-field retrieval and open-vocabulary detection described in Sec.~\ref{lab:hl} and Fig.~\ref{fig:hl}, this indicates that the full hybrid localization pipeline is beneficial for robustly grounding language queries in complex, dynamic scenes.

\section{Conclusions}
\ourname{} is a closed-loop mobile manipulation framework for dynamic, previously unseen indoor environments without a pre-built map.
By coupling a LiDAR-inertial-visual SLAM backend with a dynamic spatio-semantic voxel memory and Redundancy-Aware Memory Pruning, \ourname{} maintains a compact and temporally consistent 3D representation under frequent scene changes.
On top of this representation, the hybrid localization module, value-guided exploration/navigation scheme, and grasping-placement stack jointly enable language-conditioned target acquisition, reacquisition, and manipulation in cluttered real-world settings.
Real-robot experiments across four dynamic indoor laboratory scenes show repeated empirical gains over a strong voxel-map baseline in Navigation, Pickup, Place, and Long-term success rates, while maintaining bounded memory usage and online memory-update cost.
The current system still has several limitations. Its modular and heuristic design may accumulate errors during long-horizon execution, and its manipulation capability is mainly limited to pick-and-place tasks rather than contact-rich or articulated-object interactions.
Future work will explore active re-sensing, task rollback, tactile/force-aware manipulation, and deployment in larger cross-room environments toward long-term autonomy.

\bibliographystyle{IEEEtran}
\bibliography{IEEEabrv,reference}

\vfill

\end{document}